\documentclass{article}

\usepackage{arxiv}

\usepackage[utf8]{inputenc} 
\usepackage[T1]{fontenc}    
\usepackage{hyperref}       
\usepackage{url}            
\usepackage{booktabs}       
\usepackage{amsfonts}       
\usepackage{nicefrac}       
\usepackage{microtype}      
\usepackage{lipsum}

\usepackage{cite}
\usepackage{amsmath,amssymb,amsfonts}
\usepackage{graphicx}
\usepackage{algorithm}
\usepackage[noend]{algpseudocode}
\usepackage{hyperref}
\usepackage{textcomp}
\usepackage{threeparttablex} 

\usepackage{mathtools}
\usepackage{caption}
\usepackage{float}
\usepackage{stmaryrd}
\usepackage{epstopdf}
\usepackage{multirow}
\usepackage{pifont}
\usepackage{caption}
\usepackage{subcaption}
\usepackage{xcolor}

\newcommand{\norm}[1]{\left\lVert#1\right\rVert}

\newcommand{\R}{{\mathbb{R}}}

\newcommand{\B}{{\mathbb B}}
\newcommand{\cen}{{\mathbf{c}}}
\newcommand{\rad}{{\mathbf{r}}}
\newcommand{\ybo}{{\mathbf{y}}}
\newcommand{\ex}{\mathsf{e}}

\newcommand{\X}{{\mathbf{X}}}

\newcommand{\Y}{{\mathbf{Y}}}
\newcommand{\T}{{\mathbf{T}}}
\newcommand{\So}{{\mathbf{S}}}

\newcommand{\U}{{\mathbf{U}}}

\newtheorem{theorem}{Theorem}[section]

\newtheorem{assumption}{Assumption}

\newtheorem{definition}[theorem]{Definition}
\newtheorem{lemma}[theorem]{Lemma}
\newtheorem{remark}[theorem]{Remark}
\newtheorem{problem}[theorem]{Problem}
\newenvironment{proof}{\paragraph{Proof:}}{\hfill$\square$}

\title{Learning Spatiotemporal Tubes for Temporal Reach-Avoid-Stay Tasks using Physics-Informed Neural Networks
\thanks{ This work was supported in part by the ARTPARK. The work of Ratnangshu Das is supported by the Prime Minister’s Research Fellowship from the Ministry of Education, Government of India.}
}

\author{
 Ahan Basu \\
  Centre for Cyber-Physical Systems\\
  Indian Institute of Science, Bengaluru, India\\
  \texttt{ahanbasu@iisc.ac.in} \\
   \And
 Ratnangshu Das \\
   Centre for Cyber-Physical Systems\\
  Indian Institute of Science, Bengaluru, India\\
  \texttt{ratnangshud@iisc.ac.in} \\
  \And
 Pushpak Jagtap \\
  Centre for Cyber-Physical Systems\\
  Indian Institute of Science, Bengaluru, India\\
  \texttt{pushpak@iisc.ac.in} \\
}

\begin{document}

\maketitle

\begin{abstract}
    This paper presents a Spatiotemporal Tube (STT)-based control framework for general control-affine MIMO nonlinear pure-feedback systems with unknown dynamics to satisfy prescribed time reach-avoid-stay tasks under external disturbances. The STT is defined as a time-varying ball, whose center and radius are jointly approximated by a Physics-Informed Neural Network (PINN). The constraints governing the STT are first formulated as loss functions of the PINN, and a training algorithm is proposed to minimize the overall violation. The PINN being trained on certain collocation points, we propose a Lipschitz-based validity condition to formally verify that the learned PINN satisfies the conditions over the continuous time horizon. Building on the learned STT representation, an approximation-free closed-form controller is defined to guarantee satisfaction of the T-RAS specification. Finally, the effectiveness and scalability of the framework are validated through two case studies involving a mobile robot and an aerial vehicle navigating through cluttered environments.
\end{abstract}

\section{Introduction}

Autonomous systems have been considered as an important aspect of robotics and control systems due to their wide range of applications in safety-critical applications such as self-driving cars, warehouse automation, and unmanned aerial vehicles. These applications demand the safe and efficient execution of complex task specifications, often represented through high-level temporal logic formulations \cite{jagtap2020formal, withSTL}. The Reach-Avoid-Stay (RAS) specification serves as a building block for constructing these types of specifications \cite{RAS}. Solving such specifications is even more challenging when the system dynamics are unknown and the environment is populated with dynamic obstacles. This underlines the importance of the study of developing safe and reliable control strategies.

Several approaches have been proposed in the literature for navigation in dynamic environments, such as artificial potential field (APF) \cite{APF_global}, dynamic window approach \cite{fox2002dynamic}. Similarly, path planning techniques such as graph-based methods (A*, Dijkstra)\cite{Path_Planning_review} or sampling-based methods (PRM, RRT) \cite{sampling_review} have also been successful in guiding the agents towards reaching the goal or avoiding obstacles. However, these methods generally lack formal safety guarantees, often require separate trajectory-tracking controllers, and, crucially, do not enforce prescribed-time performance, which is essential for time-critical missions.

Control Barrier Functions (CBF) \cite{CBF} have been a promising tool in recent years in ensuring safety-critical control under the assumption of a known dynamical model. However, their performance largely depends on the chosen barrier function, and the quadratic programming (QP)-based synthesis often becomes computationally expensive for high-dimensional systems. As an alternative, \cite{Funnel_2} introduced funnel-based methods to ensure reachability and tracking, while enforcing prescribed-time performance \cite{das2024funnel}.

\cite{das2024prescribed} introduced the Spatiotemporal Tubes (STT) framework that offers a time-varying safe region in the state-space to ensure satisfaction of prescribed-time reach-avoid specifications. This has been further extended to handle more complex specifications \cite{STT_STL} and multi-agent systems \cite{STT_multi, upadhyay2025incorporating}. While the approach in \cite{das2024prescribed} uses circumvent functions that abruptly change the tube geometry, resulting in high control effort, the sampling-based counterpart \cite{das2025spatiotemporal} provides smooth tubes that can handle time-varying unsafe sets. However, the sampling-based method for constructing the STTs is highly sensitive to the choice of the basis functions used in the tube design and suffers from increased computational complexity when higher-degree polynomials are used.

Neural network-based methods have gained significant attention for controller synthesis and ensuring safety specifications. Leveraging the universal approximation property of neural networks, CBFs combined with safety controllers have been developed to ensure system safety \cite{Neural_CBF, tayal2024learning}. However, the Artificial Neural Networks (ANNs) based approaches often struggle to enforce boundary conditions. To address this, the Physics-informed Neural Networks (PINNs) have been introduced that incorporate system dynamics and boundary conditions directly into the loss functions to ensure physical consistency. PINN has been employed as a PDE solver to approximate Zubov-type PDE that satisfies Lyapunov or barrier conditions \cite{liu2025physics, agrawal2025neural}. However, these methods still require extensive data samples collected from the state-space, thereby requiring a black-box or simulator model to generate data for training these neural networks, which is difficult to obtain in real-world scenarios. Additionally it is difficult to provide formal guarantees with standard NN based methods.

In this work, we introduce the Physics-Informed Neural Spatiotemporal Tube (PINSTT) framework that constructs the STT directly from the tube dynamics without requiring state-space samples or black-box models. We formulate the STT conditions as the loss functions, with the initial and target sets imposed as the boundary constraints. The PINSTT is trained solely using time-domain collocation points, making the approach independent of sampled states. A Lipschitz-based validity condition is proposed to ensure the STT remains safe over the continuous time horizon. We then use an approximation-free, closed-form control law to keep the system within the PINSTT, ensuring the satisfaction of the T-RAS specification. Finally, the efficacy of the proposed PINSTT approach is demonstrated through two case studies: an omnidirectional robot navigating in a dynamic environment and a quadrotor operating in a cluttered workspace.

\section{Preliminaries and Problem Formulation}
\label{sec:prelim}
\subsection{Notation}
For $a, b \in \mathbb{N}$ with $a \leq b$, the closed interval in $\mathbb{N}$ is denoted as $[a; b]$. 
The set difference between two sets $\mathcal{A}$ and $\mathcal{B}$ is defined as $\mathcal{A}\setminus \mathcal{B}$.
A ball centered at $\cen \in \mathbb{R}^n$ with radius $\rad \in \mathbb{R}^+$ is defined as $\B(\cen, \rad) := \{ x \in \mathbb{R}^n \mid \|x - \cen\| \leq \rad \}$.
We define the point-to-set distance as $\mathbf{d}(x,\mathcal{A}) = \inf_{z \in \mathcal{A}} \lVert x - z \rVert$.
All other notation in this paper follows standard mathematical conventions.

\subsection{System Definition}
Consider a class of control-affine MIMO nonlinear pure-feedback systems $\mathcal{S}$ characterized by the following dynamics:
\begin{align} \label{eqn:sysdyn}
    &\dot{x}_i(t) = f_i(z_i(t)) + g_i(z_i(t))x_{i+1}(t) + w_i(t), i\in [1;N-1], \notag\\
    &\dot{x}_{N}(t) = f_N(z_N(t)) + g_N(z_N(t))u(t) + w_N(t), \\
    &y(t) = x_1(t), \nonumber
\end{align}
where for $t\in\R^+_0$ and $i\in[1;N]$,
\begin{itemize}
    \item $x_i(t) = [x_{i,1}(t), \ldots, x_{i,n}(t)]^\top \in {\X}_i \subset \mathbb{R}^{n}$ is the state,
    \item $z_i(t) := [x_1^\top(t),...,x_i^\top(t)]^\top \in \overline{\X}_i = \prod_{j=1}^i \X_j \subset \mathbb{R}^{ni} $,
    \item $u(t) \in \mathbb{R}^n$ is control input vector,
    \item $w_i(t) \in \mathbf{W} \subset \R^n$ is unknown bounded disturbance,
    \item $y(t) = [x_{1,1}(t), \ldots, x_{1,n}(t)]\in \Y=\X_1$ is the output.
\end{itemize}

The functions $f_i: \overline{\X}_i \rightarrow \mathbb{R}^n$, $g_i: \overline{\X}_i \rightarrow \mathbb{R}^{n \times n}, i \in [1;N]$, follows the Assumptions \ref{assum:lip} and \ref{assum:pd}.

\begin{assumption}\label{assum:lip}
    For all $i \in [1;N]$, functions $f_i$ and $g_i$ are unknown and locally Lipschitz.         
\end{assumption}
\begin{assumption}\cite{PPC1} \label{assum:pd}
    The \\matrix $g_{i,s}(z_i) := \frac{g_i(z_i)+g_i(z_i)^\top}{2}$ is uniformly sign-definite with known signs for all $z_i \in \overline{\X}_i$. Without loss of generality, we assume $g_{i,s}(z_i)$ is positive definite, i.e., there exists a constant $\underline{g_i}\in\mathbb R^+, \forall i \in [1;N]$ such that
    $$0 < \underline{g_i} \leq \lambda_{\min} (g_{i,s}(z_i)), \forall \ z_i \in \overline{\X}_i,$$
    where $\lambda_{\min}(\cdot)$ represents the smallest eigenvalue of the matrix.
\end{assumption}

This assumption ensures that in \eqref{eqn:sysdyn} global controllability is guaranteed, i.e., $g_{i,s}(z_i) \neq 0,$ for all $z_i \in \overline{\X}_i$.

\subsection{Problem Formulation}
Let the output of the system \eqref{eqn:sysdyn} be subjected to a temporal reach-avoid-stay (T-RAS) task, formally stated as follows:
\begin{definition}[Temporal Reach-Avoid-Stay Task]\label{def:tras}
Given a MIMO system as defined in \eqref{eqn:sysdyn} with output space $\Y = \X_1$, a prescribed time $t_c \in \R^+$, a time-varying unsafe set $\U: \R_0^+ \rightarrow \mathcal{P}(\Y)$, an initial set $\So \subset \mathbf{Y} \setminus \U(0)$, and a target set $\T \subset \mathbf{Y} \setminus \U(t_c)$, we say that the output of the system satisfies the T-RAS specifications if $y(0) \in \So$, $y(t_c) \in \T$ and for all $s \in [0,t_c], y(s) \in \mathbf{Y} \setminus \U(s)$. 
\end{definition}

\begin{problem} \label{problem:control}
Given the system in \eqref{eqn:sysdyn} satisfying Assumptions \ref{assum:lip} and \ref{assum:pd}, we aim to design an \textit{approximation-free}, \textit{closed-form} control law $u(t)$ that ensures the output $y(t)$ satisfies the T-RAS task defined in Definition \ref{def:tras}.
\end{problem}

To solve the problem, we leverage the Spatiotemporal tubes (STT) defined next.
\begin{definition}[Spatiotemporal Tubes for T-RAS tasks]\label{def:stt}
    Given a T-RAS task in Definition \ref{def:tras}, the time-varying ball $\Gamma\big(t \big):=\{x \in \R^n | \lVert x - \cen(t) \rVert \leq \rad(t)\}$, where $\rad: \R_0^+ \rightarrow \R^+$ and $\cen(t):=[\cen_1(t), \ldots, \cen_n(t)]^\top$ (with $\cen_i: \R_0^+ \rightarrow \R, \ i \in [1;n]$) are continuously differentiable functions, is called a Spatiotemporal Tube (STT) for T-RAS if the following conditions hold:
    \begin{subequations}\label{eq:stt}
    \begin{align}
        & \rad(t) > 0, \ \forall t \in [0, t_c], \\
        & \Gamma\big(0 \big) \subseteq \So, \ \Gamma\big(t_c \big) \subseteq \T, \\
        & \Gamma\big(t \big) \subseteq \Y \setminus \U(t), \forall t \in [0,t_c].
    \end{align}
    \end{subequations}
\end{definition}
This ensures that the radius of the tube is strictly positive while the tube starts from the start region and reaches the target region at time $t_c$ while avoiding the unsafe set $\U(t)$ and lying inside the output-space $\Y$.

\begin{remark}
    If one designs a control law that restricts the output trajectory of \eqref{eqn:sysdyn} within the ball, \textit{i.e.,} $\ybo \in \Gamma \big(t \big), \forall t \in [0, t_c]$ then one can ensure that the system satisfies the T-RAS specification.
\end{remark}

\section{Physics-informed Neural Spatiotemporal Tube (PINSTT)}

In this section, the main goal is to construct the STT that starts from the initial set and reaches the target set, while avoiding the unsafe set. 
\begin{lemma}
    The time-varying ball $\Gamma(t)$ characterized by center $\cen(t)$ and radius $\rad(t)$, as mentioned in Definition \ref{def:stt}, is considered a valid STT satisfying the T-RAS specification in Definition~\ref{def:tras} if the following conditions are satisfied with $\eta \leq 0$:
    \begin{subequations} \label{eq:ROP}
    \begin{align}
    & \| \cen(0) - \cen_{\So} \| = 0, \ \rad(0)  = \rad_{\So} \label{eq:start}, \\
    & \| \cen(t_c) - \cen_{\T} \| = 0, \ \rad(t_c)  = \rad_{\T} \label{eq:target}, \\
    & \forall t \in [0,t_c]: \notag \\
    & \quad \| \cen(t) - \cen_{\Y} \| + \rad(t) - \rad_{\Y} \leq \eta, \label{eq:state-space} \\
    & \quad -\rad(t) + \rad_d \leq \eta, \label{eq:radius} \\
    & \quad -\mathbf{d}(\cen(t),\U(t)) + \rad(t) \leq \eta, \label{eq:obstacle}
    \end{align}
    \end{subequations}
    where $\B(\cen_\So, \rad_\So) \subseteq \So$, $\B(\cen_{\T}, \rad_\T) \subseteq \T$, and $\B(\cen_{\Y}, \rad_\Y) \subseteq \Y$. $\rad_{d} \in \R^+$ is a user-defined lower bound on the tube radius. 
\end{lemma}
\begin{proof}
    Conditions \eqref{eq:start} and \eqref{eq:target} ensure the tube starts within the initial set and reaches the target set at the prescribed time $t_c$. For $\eta \leq 0$, condition \eqref{eq:state-space} reduces to $\lVert \cen(t) - \cen_{\Y} \rVert \leq \rad_{\Y} - \rad(t)$ implying the tube is within the boundary of the output-space $\Y$. Condition \eqref{eq:radius} with $\eta \leq 0$ ensures that the tube always has a finite radius. With $\eta \leq 0$, condition \eqref{eq:obstacle} reduces to $\rad(t) \leq \mathbf{d}(\cen(t), \U(t))$, which implies that the distance of the unsafe set from the tube center is larger than the tube radius, ensuring that the tube satisfies the avoid specification. Therefore, when the constraints in \eqref{eq:ROP} are met with $\eta \leq 0$, the obtained STT satisfies condition \eqref{eq:stt}, implying the satisfaction of the T-RAS specification.
\end{proof}

\subsection{Formal verification}

It is evident that \eqref{eq:ROP} has infinitely many constraints defined over a continuous time domain, rendering the problem intractable in its original form. To obtain a finite and computationally tractable representation, we sample $M$ points from the prescribed time horizon, denoted by $t_r$, $r\in [1;M]$. We consider a time-ball $T_r$ around each sample $t_r$ with radius $\varepsilon$, such that there exists a $t_r$ satisfying 
\begin{align}\label{eq:sample}
    | t - t_r | \leq \varepsilon, \ \forall t \in [0,t_c].
\end{align}
This ensures that the union of the time-balls forms a superset of the complete time-horizon: $\cup_{r=1}^M T_r \supset [0,t_c]$.

Our goal is to find $\Gamma(t)$ satisfying a finite number of constraints of the form of \eqref{eq:ROP} corresponding to the sampled times. However, since the structure of $\Gamma (t)$ is unknown, we can employ an artificial neural network (ANN) to approximate the time-varying ball representing the STT, leveraging the universal approximation property of neural networks. However, the major drawback of using a simple ANN is that when it is trained over some particular data points, it can neither formally ensure continuity between two points nor can it satisfy hard boundary constraints.

To overcome these limitations imposed by a simple ANN, we use a physics-informed neural network framework, namely Physics-informed Neural Spatiotemporal Tube (PINSTT). The PINSTT is denoted by 
\begin{align}\label{eqn:pinstt}
    \Gamma (t;\theta) = \B(\cen(t;\theta), \rad(t;\theta)),
\end{align} 
with center $\cen(t;\theta) = [\cen_1(t;\theta), \ldots, \cen_n(t;\theta)]^\top \in \R^n$, radius $\rad(t;\theta) \in \R$, and the trainable network parameters $\theta$.
In this formulation, the sampled time instants $t_r, r \in [1;M]$ from \eqref{eq:sample} serve as the collocation points for training, while the network outputs the curves $\cen_i(t;\theta), \ i \in [1;n]$ and $\rad(t;\theta)$. 

We now provide the following assumption and lemma to establish the main result of this subsection.
\begin{assumption} \label{assum:funlip}
    The functions $\cen(t; \theta)$ and $\rad(t; \theta)$ are Lipschitz in $t$, with constants $\mathcal{L}_{\cen}$ and $\mathcal{L}_{\rad}$, respectively.
\end{assumption}
\begin{remark}
    Note that this assumption is ensured inside the training procedure by virtue of loss functions \eqref{loss:phys_2}. In addition, this assumption is not restrictive, as Lipschitz continuity ensures a smooth tube, resulting in low control effort and avoiding the abrupt variations that require high control effort. Moreover, neural networks with slope-restricted activation functions are Lipschitz, so enforcing such regularity does not limit expressiveness while preserving the natural temporal evolution of the STT.
\end{remark}
\begin{lemma}\label{lem:point-set-dist}
If the point-to-set distances of two points $y_1$ and $y_2$ from a set $\mathcal{D}$ is defined as $\mathbf{d}(y_1,\mathcal{D})$ and $\mathbf{d}(y_2,\mathcal{D})$, then $\mathbf{d}(y_1,\mathcal{D}) - \mathbf{d}(y_2,\mathcal{D}) \leq \| y_1 - y_2 \|$.
\end{lemma}
Now we are ready to state the main theorem of this subsection.
\begin{theorem}\label{th:constr}
    The PINSTT $\Gamma(t; \theta)$ in \eqref{eqn:pinstt} trained using the sampled data points as in \eqref{eq:sample} is guaranteed to satisfy the conditions of Definition~\ref{def:stt} over the complete time-horizon $[0, t_c]$ if the following conditions are satisfied with $\hat{\eta} + \mathcal{L}\varepsilon \leq 0$:
    \begin{subequations} \label{eq:SOP}
    \begin{align}
    & \| \cen(0; \theta) - \cen_{\So} \| = 0, \ \rad(0; \theta)  = \rad_{\So} \label{eq:start_SOP}, \\
    & \| \cen(t_c; \theta) - \cen_{\T} \| = 0, \ \rad(t_c; \theta)  = \rad_{\T} \label{eq:target_SOP}, \\
    & \forall r \in [1;M], t_r \in [0,t_c]: \notag \\
    & \quad \| \cen(t_r; \theta) - \cen_{\Y} \| + \rad(t_r; \theta) - \rad_{\Y} \leq \hat{\eta} \label{eq:state_SOP}, \\
    & \quad -\rad(t_r; \theta) + \rad_d \leq \hat{\eta}, \label{eq:rad_SOP} \\
    & \quad -\mathbf{d}(\cen(t_r; \theta),\U(t_r)) + \rad(t_r; \theta) \leq \hat{\eta} \label{eq:avoid_SOP},
    \end{align}
    \end{subequations}
\end{theorem}
\begin{proof}
This proof shows that if condition $\hat{\eta} + \mathcal{L}\varepsilon \leq 0$ is met, the STT $\Gamma (t; \theta)$ satisfies Definition~\ref{def:stt}. From \eqref{eq:sample}, it is evident that for every $t \in [0, t_c]$, there exists a sampled point $t_r$ such that $|t - t_r| \leq \varepsilon$. Therefore, for all $t \in [0, t_c]$:
\begin{align*}
    \text{(i)} &-\rad(t; \theta) + \rad_d = -\rad(t_r; \theta) + \rad_d + \rad(t_r; \theta) - \rad(t; \theta) \leq \ \hat{\eta} + \mathcal{L}_\rad |t-t_s| \leq \hat{\eta} + \mathcal{L}_\rad\varepsilon \leq \hat{\eta} + \mathcal{L}\epsilon \leq 0.
\end{align*}
This implies that $\rad(t; \theta) \geq \rad_d > 0$ for all $t \in [0, t_c]$.
\begin{align*}
     \text{(ii) } &\| \cen(t; \theta) - \cen_{\Y} \| + \rad(t; \theta) - \rad_{\Y} \\
     = &\| \cen(t; \theta) - \cen_{\Y} \| - \| \cen(t_r; \theta) - \cen_{\Y} \| + \rad(t; \theta) - \rad(t_r; \theta) +\| \cen(t_r; \theta) - \cen_{\Y} \| + \rad(t_r; \theta) - \rad_{\Y} \\
     \leq & \mathcal{L}_\cen |t-t_r| + \mathcal{L}_\rad |t-t_r| + \hat{\eta} \leq \hat{\eta} + \mathcal{L}\epsilon \leq 0.
\end{align*}
This implies that $\Gamma (t; \theta) \subseteq \X$ for all $t \in [0, t_c]$.
\begin{align*}
    \text{(iii) } &- \mathbf{d}(\cen(t; \theta),\U(t)) + \rad(t; \theta) \\ 
    = &- \mathbf{d}(\cen(t; \theta),\U(t)) + \mathbf{d}(\cen(t_r; \theta),\U(t_r)) + \rad(t; \theta) - \rad(t_r; \theta) - \mathbf{d}(\cen(t_r; \theta),\U(t_r)) + \rad(t_r; \theta) \\
    \leq &\|\cen(t; \theta) - \cen(t_r; \theta)\| + \|\rad(t; \theta) - \rad(t_r; \theta)\| + \hat{\eta} \leq (\mathcal{L}_{\cen} + \mathcal{L}_{\rad})\epsilon \leq \mathcal{L}\epsilon + \hat{\eta} \leq 0
\end{align*}
This implies that $\Gamma (t; \theta) \cap \U = \emptyset$ for all $t \in [0, t_c]$. Hence, satisfying the conditions \eqref{eq:SOP} with $\hat{\eta} + \mathcal{L}\varepsilon \leq 0$ guarantees the tube $\Gamma(t; \theta)$ will satisfy the T-RAS task.
\end{proof}

\subsection{Formulation of loss functions and training procedure}

The constraints in the SOP \eqref{eq:SOP} are encoded as physics-based penalties in the loss functions of the PINN framework. While the state-space constraint \eqref{eq:state_SOP}, radius constraint \eqref{eq:rad_SOP} and obstacle avoidance constraint \eqref{eq:avoid_SOP} are considered as the physics loss of the training procedure, the start and target conditions \eqref{eq:start_SOP} and \eqref{eq:target_SOP} are enforced as boundary conditions. The overall loss function that is to be minimized is given by:
\begin{align}\label{eq:loss_tot}
    \mathbb{L}(\theta) = \mathbb{L}_{phys}(\theta) + \mathbb{L}_{bc}(\theta),
\end{align}
while the physics-based loss $\mathcal{L}_{phys}(\theta)$ and boundary loss $\mathcal{L}_{bc}(\theta)$ are the sum of the sub-loss terms formulated by the constraints. 

We define the sub-loss terms that are used to construct the physics-based loss as:
\begin{subequations}\label{loss:phys_1}
\begin{align}
    \mathbb{L}_{p,1}(\theta) &= \sum_{r=1}^M \text{ReLU}\big(\| \cen(t_r; \theta) - \cen_{\Y} \| + \rad(t_r; \theta) - \rad_{\Y} - \hat{\eta}\big), \\
    \mathbb{L}_{p,2}(\theta) &= \sum_{r=1}^M \text{ReLU} \big(-\rad(t_r; \theta) + \rad_d - \hat{\eta}\big), \\
    \mathbb{L}_{p,3}(\theta) &= \sum_{r=1}^M \text{ReLU} \big(-\mathbf{d}(\cen(t_r; \theta),\U(t_r)) + \rad(t_r; \theta) - \hat{\eta}\big), 
\end{align}
\end{subequations}
where $\text{ReLU}(x) = \max(0,x)$. We need to provide a guarantee that the solution obtained upon training the PINSTT is valid for the continuous domain by virtue of Theorem \ref{th:constr}. Therefore, it is necessary to guarantee the Lipschitz continuity of $\cen(t; \theta)$ and $\rad(t; \theta)$, as stated in Assumption \ref{assum:funlip}. We leverage the automatic differentiation property of PINNs to ensure that they remain bounded and thus satisfy the Lipschitz conditions. To achieve this, we introduce two additional sub-loss terms within the physics-based loss function, defined as follows:
\begin{subequations}\label{loss:phys_2}
\begin{align}
    \mathbb{L}_{p,4}(\theta) &= \sum_{r=1}^M \text{ReLU}(\| \dot{\cen}(t_r; \theta) \| - \mathcal{L}_c), \\
    \mathbb{L}_{p,5}(\theta) &= \sum_{r=1}^M \text{ReLU} (\| \dot{\rad}(t_r; \theta) \| - \mathcal{L}_r).
\end{align}
\end{subequations}
Therefore, the combined physics-based loss is given by:
\begin{align}\label{eq:loss_phys}
    \mathbb{L}_{phys}(\theta) = \sum_{i=1}^5 w_{p,i} \mathbb{L}_{p,i}(\theta),
\end{align}
where $w_{p,i}\geq0, i \in [1;5]$ are the weights corresponding to the sub-loss terms. Similarly, the boundary loss function is given by:
\begin{align}\label{eq:loss_bc}
    &\mathbb{L}_{bc} (\theta) = w_{b,1} \text{MSE}(\cen (0; \theta), \cen_{\So}) + w_{b,2} \text{MSE}(\rad(0; \theta), \rad_{\So}) + w_{b,3} \text{MSE}(\cen (t_c; \theta), \cen_{\T}) + w_{b,4} \text{MSE}(\rad(t_c; \theta), \rad_{\T}),
\end{align}
with $w_{b,i}\geq 0, i \in [1;4]$ are the weights corresponding to the sub-loss functions and $\text{MSE}(x,y) = \| x - y \|^2$ stands for mean-square error.

The training procedure of the PINSTT is shown in Algorithm \ref{algo:NN_training}.

\begin{algorithm}
\caption{Training of the PINSTT}
\label{algo:NN_training}
\begin{algorithmic}[1]
    \Require Time collocation data $\mathcal{T}:=\{t_r\}_{r=1}^M$, T-RAS specifications, batch size, learning rate, maximum number of epochs. 
    \Ensure $\Gamma (t; \theta)$
    \State Initialize PINN and trainable parameter $\theta$. Initialize $\hat{\eta}$ such that $\hat{\eta} = - \mathcal{L}\varepsilon$.
    \For{$i\leq Epochs$ (Training starts here)}
        \State Create batches of training data from $\mathcal{T}$
        \State Find batch losses using \eqref{eq:loss_phys} and \eqref{eq:loss_bc}.
        \State Compute total loss $\mathbb{L}(\theta)$ using \eqref{eq:loss_tot}.
        \State Update the NN parameters $\theta$ using Adam/SGD to minimize the loss.
        \If{$\mathbb{L}(\theta) \approx 0$}
            \State \textbf{break}
        \EndIf
    \EndFor
    \State \textbf{return} $\Gamma (t; \theta)$
\end{algorithmic}
\end{algorithm}
\begin{remark}
    In practice, due to data complexity or numerical precision, the loss may not converge exactly to zero. Therefore, we assume a tolerance of the order $10^{-6}$ to $10^{-4}$ for $\mathbb{L}(\theta)$ during implementation. 
\end{remark}
\begin{remark}
    The convergence of the algorithm can be improved by reducing the discretization parameter $\varepsilon$ [\cite{zhao2021learning}], tuning neural network hyperparameters (e.g., architecture, learning rate) [\cite{nn_lr}], or revising the algorithm’s initial hyperparameters.
\end{remark}
\section{Controller Synthesis}
In this section, we derive a closed-form, approximation-free control law to constrain the system output of the agent within the STT. We leverage the lower triangular structure in system dynamics for each agent in \eqref{eqn:sysdyn}. We first synthesize the control input $r_2$ to enforce the tube constraints on the system output. Then, we use the idea from \cite{das2025spatiotemporal} to recursively define intermediate reference signals $r_{z+1}$ such that each state $x_z$ tracks $r_z$ for all $z \in [2;N]$ with $u = r_{N+1}$ as the final control input. The steps of the control design are as follows:

\textbf{Stage $1$}: Given the STT $\Gamma (t; \theta)$ of Definition \ref{def:stt}, define the normalized error $e_1(x_1,t)$ and the transformed error $\varepsilon_1(x_1,t)$ as
\begin{align}
e_1(x_1,t) = \frac{\norm{x_1(t) - \cen(t)}}{\rad(t)},
\varepsilon_1(x_1,t) = \ln\left( \frac{1 + e_1(x_1,t)}{1 - e_1(x_1,t)} \right).\nonumber
\end{align}
The intermediate control input $r_2(x_1,t)$ is then given by 
\begin{equation}
    r_2(x_1,t) = -\kappa_1 \varepsilon_1(x_1,t) \big( x_1(t)-\cen(t) \big), \kappa_1 \in \R^+.
\end{equation}

\textbf{Stage $k$} ($k \in [2;N]$): 
To ensure $x_k$ tracks the reference signal $r_k$ from Stage $k-1$, we define a time-varying bound: $\gamma_{k,i}(t) = (p_{k,i} - q_{k,i})\ex^{-\mu_{k,i}t} + q_{k,i}$, and enforce,
    $-\gamma_{k,i}(t) \leq (x_{k,i}-r_{k,i}) \leq \gamma_{k,i}(t), \forall (t,i) \in \R_0^+ \times [1;n],$
where, $\mu_{k,i} \in \R_0^+$, and $p_{k,i}> q_{k,i} \in \R^+$ are chosen such that $|x_{k,i}(0) - r_{k,i}(0)| \leq p_{k,i}$.
Now, define the normalized error $e_k(x_{k},t)$, the transformed error $\varepsilon_k(x_{k},t)$ and $\xi_k(x_{k},t)$ as follows
\begin{subequations} \label{eq:stage k}
    \begin{align}
    e_k(x_{k},t) &= [e_{k,1}(x_{k,1},t), \ldots, e_{k,n}(x_{k,n},t)]^\top = (\textsf{diag}(\gamma_{k,1}(t),\ldots,\gamma_{k,n}(t)))^{-1} \left(x_{k} - r_k \right), \\ 
    \varepsilon_k(x_{k},t) &= \big[\ln\left(\frac{1+e_{k,1}(x_{k,1},t)}{1-e_{k,1}(x_{k,1},t)}\right), \ldots, \ln\left(\frac{1+e_{k,n}(x_{k,n},t)}{1-e_{k,n}(x_{k,n},t)}\right) \big]^\top, \\
    \xi_k(x_{k},t) &= 4 \big(\textsf{diag}(\gamma_{k,1}(t),\ldots,\gamma_{k,n}(t)) \big)^{-1} (I_n - \textsf{diag}(e_k \circ e_k))^{-1}.
\end{align}
\end{subequations}
The next intermediate control input $r_{k+1}(\overline{x}_{k},t)$ is then:
\begin{equation*}
    r_{k+1}(\overline{x}_{k},t) = - \kappa_k\xi_k(x_{k},t)\varepsilon_k(x_{k},t), \kappa_k \in \R^+.
\end{equation*}
At stage $N$, this intermediate input is the actual control input:
\begin{equation*}
    u(\overline{x}_{N},t) = - \kappa_N\xi_N(x_{N},t)\varepsilon_N(x_{N},t), \kappa_N \in \R^+.
\end{equation*}
We now state the main theorem guaranteeing that this controller enforces the desired T-RAS behavior.
\begin{theorem} \label{theorem_ras}
    Consider the nonlinear MIMO system \eqref{eqn:sysdyn} under Assumptions \ref{assum:lip} - \ref{assum:pd}, with a T-RAS task (Definition~\ref{def:tras}), and the STT $\Gamma (t; \theta)$ from Definition \ref{def:stt}. If the initial output is within the STT: $y(0) \in \Gamma (0; \theta)$, then the controller
    \begin{align}\label{eqn:Control_ras}
        r_2(x_1,t) &= -\kappa_1 \varepsilon_1(x_1,t) \left( x_1(t)-\cen(t) \right), \kappa_1 \in \R^+, \notag \\
        r_{k+1}(\overline{x}_{k},t) &= - \kappa_k\xi_k(x_{k},t)\varepsilon_k(x_{k},t), k \in [2;N-1], \notag \\
        u(\overline{x}_{N},t) &= - \kappa_N\xi_N(x_{N},t)\varepsilon_N(x_{N},t),
    \end{align}    
    ensure that the system output remains within the STT: 
    $$y(t) = x_1(t) \in \Gamma (t; \theta), \forall t \in [0, t_c],$$
    thereby satisfying the desired T-RAS specification.
\end{theorem}
\begin{proof}
    The proof is similar to the proof of \cite{upadhyay2025incorporating} and hence omitted due to space constraints.
\end{proof}


\section{Case Study}
To validate the effectiveness of the proposed approach, we present two different case studies: a 2D omnidirectional robot and a 3D UAV. For both case studies, we present two scenarios comprising both static and dynamic obstacles. All experiments were conducted in PyTorch (Python 3.10) on a Windows machine with an Intel Core i7-14700 CPU, 32 GB RAM, and an NVIDIA GeForce RTX 3080 Ti GPU. 

\subsection{Omnidirectional Robot}

\begin{figure*}[h!]
    \centering
    \includegraphics[width=0.95\textwidth]{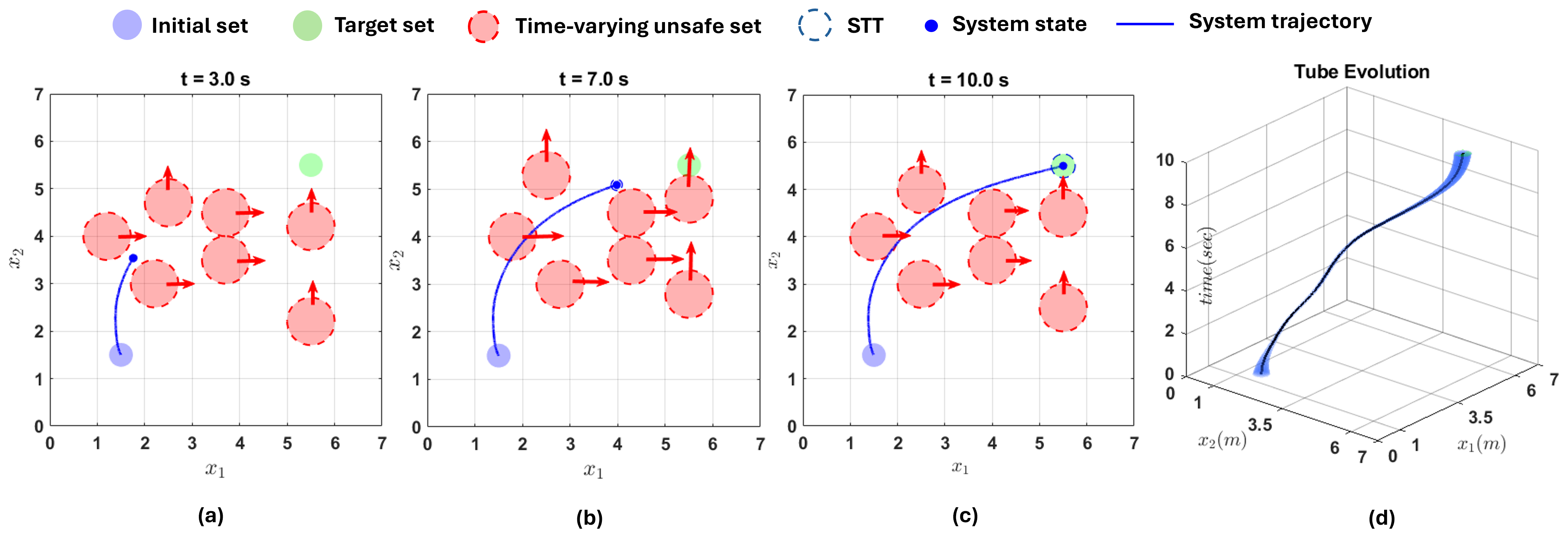}
    \vspace{-0.25cm}
    \caption{The constructed STT and the omnibot trajectory in a dynamic environment with time-varying obstacles.}
    \label{fig:omni}
\end{figure*}

For the first case study, we consider the omnidirectional robot adopted from \cite{jagtap2024controller} given by
\begin{align}
    \begin{bmatrix}
        \dot{x}_1 \\ \dot{x}_2 \\ \dot{x}_3
    \end{bmatrix}
    = 
    \begin{bmatrix}
        \cos{x_3} & -\sin{x_3} & 0 \\ \sin{x_3} & \cos{x_3} & 0 \\ 0 & 0 & 1
    \end{bmatrix}
    \begin{bmatrix}
        v_1 \\ v_2 \\ \omega
    \end{bmatrix} + w(t),
\end{align}
where the state vector $[x_1, x_2, x_3]^\top$ captures the pose of the robot, $[v_1, v_2, \omega]^\top$ is the input velocity vector in the frame of the robot. The robot starts from the initial set $\So = \mathbb{B}([1.5, 1.5]^\top, 0.25) $ and has to reach the target zone $\T = \mathbb{B}([5.5, 5.5]^\top, 0.25) $ within a prescribed time of $t_c = 10$ sec while the state-space is constrained within the region $[0,7] \times[0,7]$. The robot navigates through a cluttered environment with dynamically oscillating unsafe regions. The obstacles with the horizontal arrows oscillate horizontally, while the ones with the vertical arrows oscillate vertically. Figure \ref{fig:omni}(a)-(c) shows the position of the robot at different time stamps, while Figure \ref{fig:omni}(d) presents the evolution of the STT over the complete time-horizon. The offline computation time to obtain the STT through the training of PINN is 7.854 seconds, while the online control synthesis time is 0.008 seconds.

\subsection{Quadrotor}
\begin{figure}[h]
    \centering
    \includegraphics[width=0.6\linewidth]{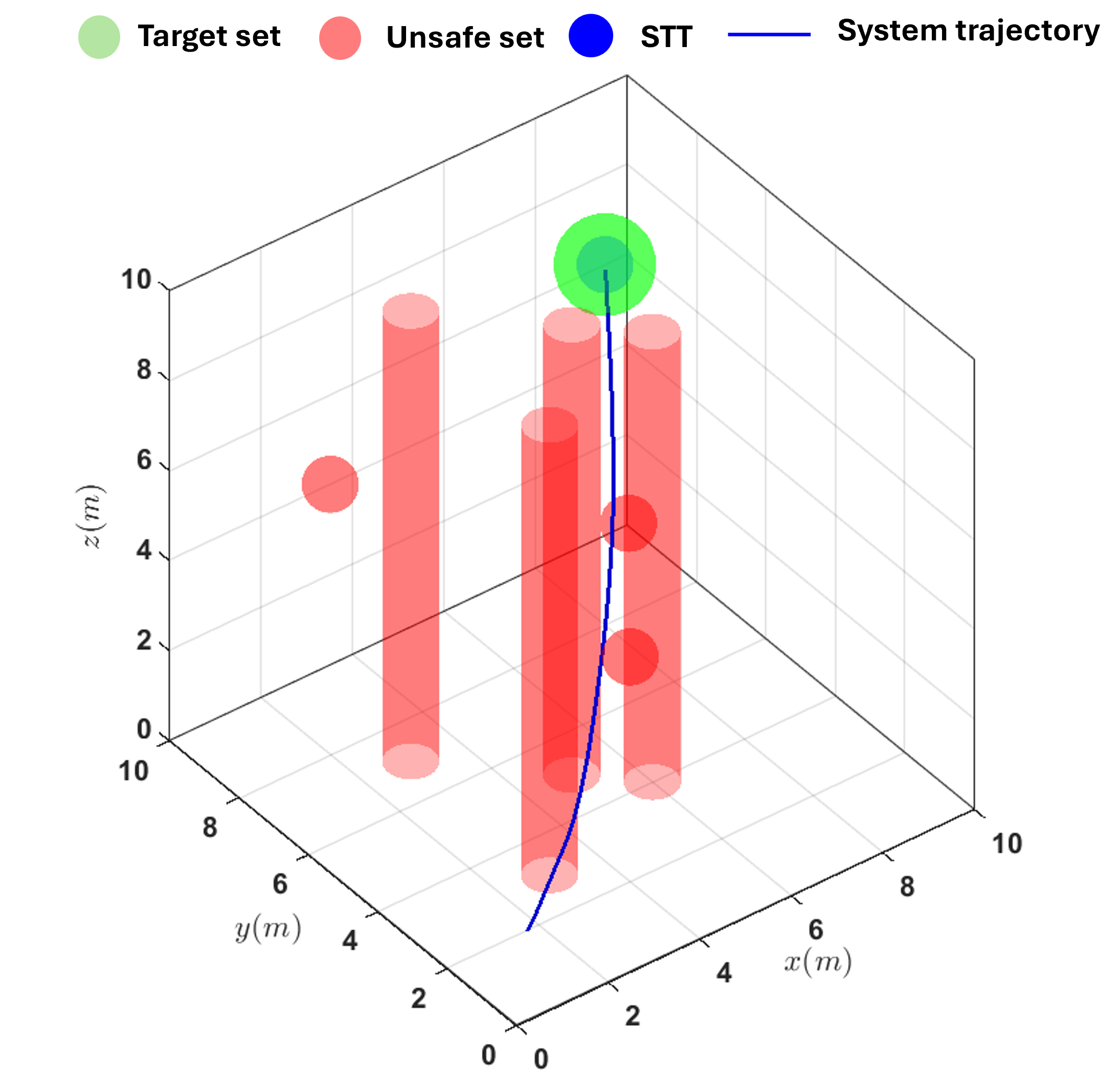}
    \caption{The drone trajectory in a static environment with static obstacles.}
    \label{fig:drone}
\end{figure}

We further evaluate the framework on a Quadrotor operating in a 3D environment with second-order dynamics adapted from \cite{APF_drone} given by
\begin{align}
    \left[\dot{x} \ \dot{y} \ \dot{z} \right]^\top    &= \left[v_x\ v_y \ v_z \right]^\top + w_1(t), \\
    \left[v_x \ v_y \ v_z \right]^\top &= \left[u_x \ u_y \ u_z \right]^\top + w_2(t),
\end{align}
where $[x,y,z]^\top$ captures the position, $[v_x,v_y,v_z]^\top$ is the velocity and $[u_x,u_y,u_z]^\top$ is the control input of the drone.
The system starts from $\So = \mathcal{B}([1,1,1]^\top, 0.8)$ and must reach the target $\T = \mathcal{B}([8,8,8]^\top, 0.8)$ within a prescribed time of $t_c = 10$ sec, while avoiding multiple static obstacles within the state-space $[0,10] \times [0,10] \times [0,10] $. Figure \ref{fig:drone} shows the quadrotor's trajectory avoiding all the unsafe regions and reaching the target. The offline computation time to obtain the STT through the training of PINN is 3.252 seconds, while the online control synthesis time is 0.021 seconds.

\section{Conclusion and future work}

This work presents a Physics-Informed Neural Network (PINN)-based framework to synthesize STT satisfying T-RAS specifications for unknown control-affine MIMO pure feedback systems under unknown but bounded disturbances. Leveraging the universal approximation capability of neural networks, the time-varying center and radius of the tube are jointly learned through a PINN, while the Lipschitz continuity required for formal guarantees is enforced via the network’s physics-informed loss. Subsequently, a closed-form, model-free controller is used to ensure that the system trajectories remain within the STT, thereby fulfilling the T-RAS objectives. The proposed framework’s effectiveness and generalizability are demonstrated through multiple case studies.
Despite its advantages, the current formulation does not explicitly incorporate input constraints, which may limit its applicability in real-world systems and is an important direction for future research.

\bibliographystyle{plain}
\bibliography{sources}

\end{document}